\documentclass{article}
\usepackage{ijcai26}

\usepackage{times}
\usepackage{soul}
\usepackage{url}
\usepackage[hidelinks]{hyperref}
\usepackage[utf8]{inputenc}
\usepackage[small]{caption}
\usepackage{graphicx}
\usepackage{amsmath,amsfonts}
\usepackage{amsthm}
\usepackage{booktabs}
\usepackage{algorithm}
\usepackage{algorithmic}
\usepackage[switch]{lineno}

\usepackage{caption}
\usepackage{subcaption}
\usepackage{booktabs}
\usepackage{xcolor}

\definecolor{mybg}{rgb}{0.98,0.98,0.98}  % 柔灰背景
\definecolor{myframe}{rgb}{0.4,0.4,0.4}  % 深灰边框
\definecolor{mygreen}{rgb}{0,0.5,0}  % 深绿色

\usepackage{newfloat}
\usepackage{listings}
\DeclareCaptionStyle{ruled}{labelfont=normalfont,labelsep=colon,strut=off} % DO NOT CHANGE THIS
\floatstyle{ruled}
\newfloat{listing}{tb}{lst}{}
\floatname{listing}{Listing}

\lstdefinestyle{fancylisting}{
    language=Python,
    basicstyle=\ttfamily\small\color{black},
    keywordstyle=\color{blue},
    stringstyle=\color{teal},
    commentstyle=\color{gray}\itshape,
    numbers=none,
    numberstyle=\tiny\color{gray},
    stepnumber=1,
    numbersep=8pt,
    backgroundcolor=\color[RGB]{250,250,250},
    frame=single,
    rulecolor=\color{gray},
    breaklines=true,
    breakatwhitespace=true,
    showstringspaces=false,
    captionpos=b,
    tabsize=4,
    keepspaces=true,
    aboveskip=1em,   % 上方留白
    belowskip=1em,   % 下方留白
    framexleftmargin=0.3em,
    framexrightmargin=0.2em,
    framextopmargin=0.2em,
    framexbottommargin=0.2em,
}

\definecolor{modelGPT}{HTML}{4E79A7}
\definecolor{modelDoubao}{HTML}{59A14F}
\definecolor{modelDeepSeek}{HTML}{F28E2B}
\definecolor{modelClaude}{HTML}{B07AA1}
\definecolor{modelMistral}{HTML}{76B7B2}

\newcommand{\dvblegendline}[1]{\raisebox{0.35ex}{\textcolor{#1}{\rule{1.35em}{1.0pt}}}} \newcommand{\dvblegenditem}[2]{\dvblegendline{#1}\hspace{0.25em}#2}

\title{DiverValue-Bench: A Benchmark and Fine-Tuning Framework for Aligning Large Language Models with Diverse Human Values}

\author{
Yao Liang$^{1,5}$\and
Dongcheng Zhao$^1$\and
Feifei Zhao$^{1,2,5}$\thanks{Corresponding author.}\and
Guobin Shen$^{1,6}$\and
Yuwei Wang$^1$\and
Dongqi Liang$^{1,5}$\And
Yi Zeng$^{1,2,3,4}$\footnotemark[1]\\
\affiliations
$^1$Brain-inspired Cognitive AI Lab, Institute of Automation, Chinese Academy of Sciences\\
$^2$Beijing Key Laboratory of Safe AI and Superalignment\\
$^3$Gaoling School of AI, Renmin University of China\\
$^4$Beijing Institute of AI Safety and Governance (Beijing-AISI)\\
$^5$School of Artificial Intelligence, University of Chinese Academy of Sciences\\
$^6$School of Future Technology, University of Chinese Academy of Sciences\\
\emails
liangyao2023@ia.ac.cn,
zhaofeifei2014@ia.ac.cn,
yi.zeng@ruc.edu.cn
}

\begin{document}

\maketitle

\begingroup
\renewcommand{\thefootnote}{}
\footnotetext{Open-science materials are publicly available at: \url{https://github.com/YaoLiang2023/DiverValue-Bench}.}
\endgroup

\begin{abstract}
Aligning large language models (LLMs) with diverse human values is essential for safe and effective deployment, yet existing benchmarks often overlook cultural and demographic variation. We introduce DiverValue-Bench, a population-aware benchmark for evaluating multi-dimensional value alignment across 74 countries/regions. It contains 23,763 quality-controlled instances derived from PRISM user feedback and audited through large-scale human validation, with fine-grained value labels, personalized questions, contrastive reference answers, and rich demographic metadata. Using DiverValue-Bench, we evaluate representative LLMs and reveal substantial geographic and demographic disparities that are masked by aggregate performance. We further show that lightweight preference-based fine-tuning with Low-Rank Adaptation (LoRA) and Direct Preference Optimization (DPO) substantially improves in-domain value alignment while yielding consistent out-of-domain gains. These results highlight the need for population-aware alignment evaluation and demonstrate the utility of DiverValue-Bench as a practical foundation for global alignment, personalized value modeling, and equitable AI development.
\end{abstract}

\section{Introduction}
Large language models (LLMs) have demonstrated impressive capabilities in natural language understanding and generation, powering a wide range of real-world applications across education, healthcare, law, and creative industries~\cite{ouyang2022training,ding2023parameter,ziems-etal-2023-large,achiam2023gpt,touvron2023llama2,dubey2024llama}. However, as LLMs increasingly interact with diverse user populations, a critical challenge emerges: ensuring that their outputs align with the ethical norms, social values, and personal preferences of humans—a problem commonly referred to as the value alignment challenge~\cite{li2023self,xu2023some,shen2023large,lee2024llm2llm,asai2024self,yu2024self,azar2024general}.

Value alignment aims to ensure that LLMs produce content that is ethically appropriate, socially acceptable, and individually respectful. Existing approaches such as reinforcement learning with human feedback (RLHF)~\cite{ziegler2019fine,ouyang2022training} and direct preference optimization (DPO)~\cite{rafailov2023direct} have improved alignment performance, particularly in high-stakes and safety-sensitive scenarios. Nevertheless, these methods often rely on limited or homogeneous datasets, which hinders their generalization to personalized, cross-cultural, and multi-value contexts~\cite{kasirzadeh2023conversation,kirk2023personalisation,freedman2023active,motoki2024more,rozado2024political}.

Recent work has explored emerging directions such as personalized alignment, value pluralism, and cross-cultural evaluation. For example, datasets like PRISM~\cite{kirk2024prism} and PERSONA~\cite{castricato2024persona} incorporate multi-value feedback to better reflect individual preferences, while the Value Kaleidoscope~\cite{sorensen2024value} project highlights the importance of evaluating alignment robustness under culturally divergent and morally ambiguous conditions. These efforts mark important progress, but significant limitations remain: 1) current datasets rarely contain rich, fine-grained user profiles (e.g., age, gender, education), making it difficult to simulate realistic and diverse alignment scenarios; 2) many studies implicitly assume a universal ethical framework, overlooking the inherent plurality and conflict of values across populations, societies, and time periods. Mainstream benchmarks are heavily skewed toward Western-centric value systems, inadvertently marginalizing non-mainstream or alternative viewpoints~\cite{roche2023ethics,friederich2024symbiosis}; 3) existing evaluation methods typically rely on single-dimensional consistency metrics, failing to capture structured variations across cultures, identity groups, or ethical dimensions. These limitations collectively impede our ability to assess and improve LLM alignment in complex, heterogeneous settings.

To address the aforementioned challenges, we propose DiverValue-Bench, a benchmark for multi-value preference alignment. It comprises 23,763 quality-controlled instances across 7 value dimensions, 1,396 users, and 74 countries/regions, with a blind human audit covering 6,406 sampled instances. Each user is annotated with detailed demographic information—including age, gender, education level, employment status, language proficiency, and marital status—enabling fine-grained and context-aware analysis of alignment behaviors. Building on PRISM user feedback, we automatically generate profile-conditioned Q\&A pairs with consistency filtering and targeted manual audits to improve quality.

The construction of DiverValue-Bench serves as a foundation for systematic, large-scale evaluation of alignment performance under heterogeneous human value systems. Building on this dataset, we further develop a structured evaluation framework that facilitates comparative analysis across a range of widely used language models under varied demographic and ethical contexts. Empirical results derived from DiverValue-Bench reveal substantial disparities in alignment performance across different population groups, highlighting key limitations of current alignment strategies in globally deployed systems. Moreover, we demonstrate that lightweight adaptation techniques guided by value-diverse supervision can effectively enhance model robustness and cultural sensitivity, underscoring the practical utility of DiverValue-Bench in advancing the development of more inclusive and globally aligned language models. 

Our key contributions are as follows:

\begin{itemize}
    \item \textbf{Multi-dimensional contrastive benchmark.}
    We introduce DiverValue-Bench ($23{,}763$ instances; $1{,}396$ users; $74$ countries/regions), which pairs each question with a contrastive reference pair (aligned vs.\ misaligned) under a 7-D value-preference schema and detailed user profiles, enabling population- and subgroup-level audits of value-pluralistic alignment.

    \item \textbf{Structured cross-cultural evaluation.} We propose a structured evaluation framework for fine-grained country/region and subgroup analysis, revealing systematic alignment disparities that aggregate scores can mask.

    \item \textbf{Lightweight preference-based alignment.}
    Using DiverValue-Bench preference pairs as supervision, we show that lightweight alignment tuning (LoRA/DPO) on LLaMA-2 and Qwen substantially improves preference alignment (OPA $\sim27\% \rightarrow 78$--$89\%$) and yields consistent out-of-domain gains, demonstrating practical utility for culturally adaptive value alignment.
\end{itemize}

\section{Related Work}

Recent advances in value alignment for LLMs have predominantly leveraged RLHF and DPO to align model behavior with normative ethical standards. While these approaches have achieved notable success, they often assume a universal value framework, neglecting the complexities introduced by value pluralism, individual differences, and cross-cultural variability~\cite{chakraborty2024maxmin,wang2024map,jang2023personalized,yang2024rewards}. To address these limitations, several recent datasets have begun to explicitly encode diverse human value systems, offering new opportunities for nuanced alignment research~\cite{poddar2024personalizing,peschl2021moral,rame2023rewarded}.

% \paragraph{Value Pluralism.}  
\textbf{Value pluralism} emphasizes the need to respect and represent the diversity of human values rather than enforcing a singular normative standard. For example, the PERSONA dataset~\cite{castricato2024persona} constructs synthetic user personas encompassing varied demographic attributes and value stances, enabling the evaluation and improvement of pluralistic alignment strategies. Similarly, Value Kaleidoscope~\cite{sorensen2024value} introduces the ValuePrism dataset, which captures conflicting principles, rights, and duties to simulate morally complex decision-making scenarios. These resources support the development of LLMs capable of navigating value conflicts in real-world settings.

% \paragraph{Personalized Alignment.}  
\textbf{Personalized alignment} focuses on tailoring model outputs to reflect individual user preferences, thereby improving perceived helpfulness and satisfaction. Early approaches often relied on single-dimensional preference modeling, which fails to capture the complexity of real-world value systems. PRISM~\cite{kirk2024prism} demonstrates that leveraging user profiles for personalized feedback substantially enhances alignment accuracy. Expanding on this, \citeauthor{li20251}~\shortcite{li20251} proposed multi-dimensional preference spaces derived from psychological theories to better extract and model fine-grained individual preferences.

% \paragraph{Cross-Cultural Alignment.}  
\textbf{Cross-cultural alignment} aims to ensure model generalization and fairness across global populations. PRISM provides explicit value feedback and demographic attributes, but it does not offer contrastive, preference-labeled Q\&A pairs suitable for benchmarking and preference-based fine-tuning under a unified evaluation protocol. The Moral Machine experiment~\cite{awad2018moral} provides large-scale evidence of cultural variation in ethical decision-making, revealing systematic biases and underscoring the need for culturally sensitive alignment frameworks.

% \paragraph{Limitations of Existing Work.}  
Overall, existing work has advanced pluralistic, personalized, and cross-cultural alignment, yet a gap remains between diverse feedback collection and actionable benchmarking. Many resources either provide limited demographic granularity, lack contrastive preference-labeled Q\&A pairs, or offer insufficient protocols for subgroup-level evaluation and preference-based tuning. DiverValue-Bench fills this gap by combining multi-dimensional value labels, fine-grained user profiles, and aligned/misaligned reference pairs within a unified framework for population-aware alignment evaluation and adaptation.

\section{DiverValue-Bench Construction}

\begin{figure*}[t]
  \centering
  \includegraphics [width=17cm]{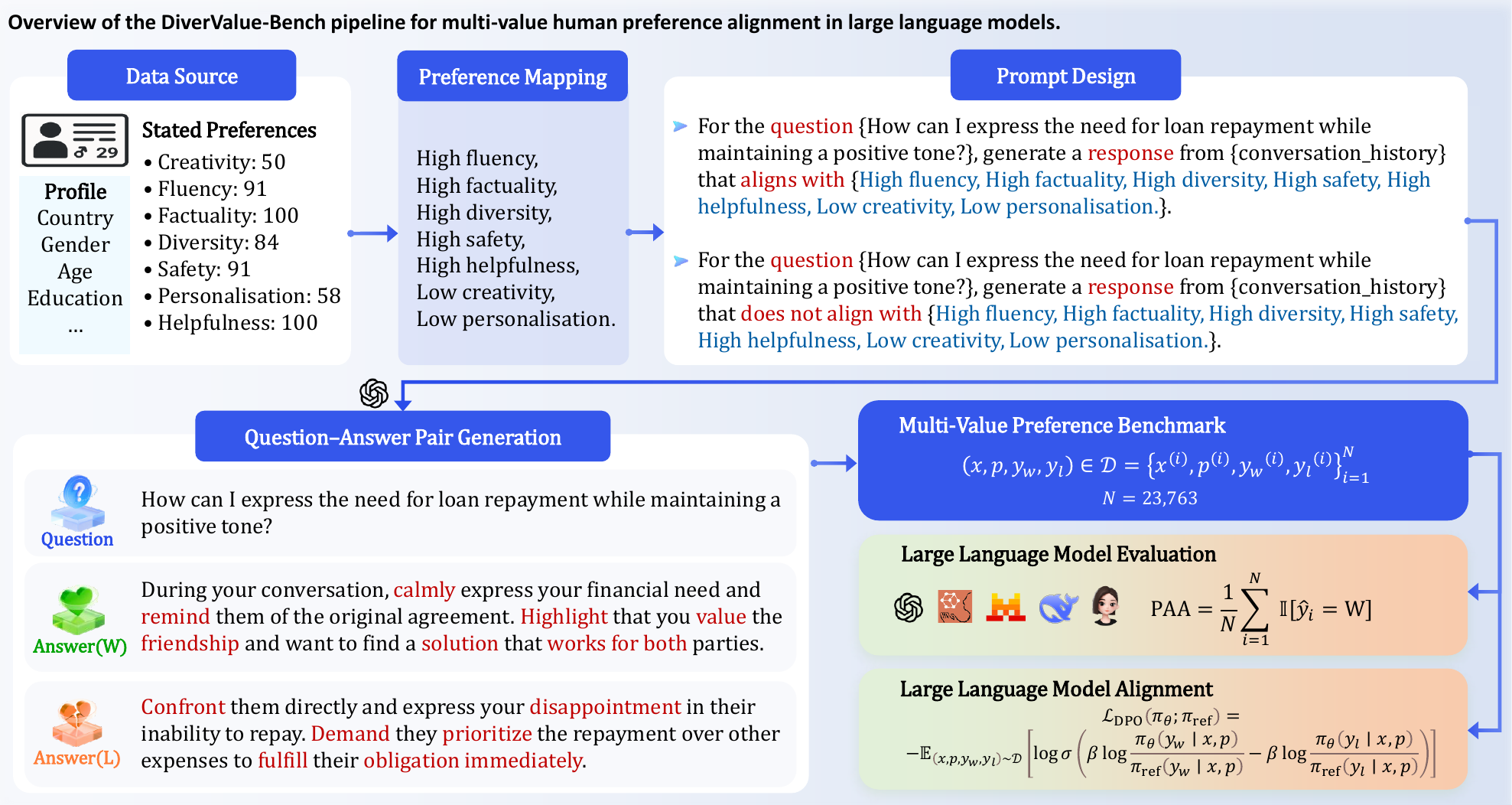}
  \caption{Overview of the DiverValue-Bench pipeline for multi-value human preference alignment in large language models. We map explicit user feedback to 7-D value-preference labels, generate contrastive Q\&A pairs $(x,p,y_w,y_l)$ (aligned $y_w$ vs.\ misaligned $y_l$) conditioned on the question $x$ and user profile $p$, and attach demographic metadata. The resulting benchmark ($N{=}23{,}763$) supports alignment evaluation via Preference Alignment Accuracy (PAA) and preference-based fine-tuning (e.g., LoRA, DPO).}
  \label{fig:DiverValueBench_main}
\end{figure*}

In this paper, we propose and construct a dataset, DiverValue-Bench, specifically designed for aligning LLMs with diverse human values. The dataset aims to address the current limitations in understanding and generating outputs aligned with diverse human values. Emphasizing high-quality and large-scale data collection, DiverValue-Bench systematically captures users' multi-dimensional value preferences and personalized requirements. The dataset construction involves three stages: (1) Value Preference Mapping; (2) Personalized Q\&A Generation; and (3) User Profile Integration. An overview of this pipeline is illustrated in Fig.~\ref{fig:DiverValueBench_main}.

\subsection{Value Preference Mapping}

We map users' explicit feedback to interpretable value-preference labels. Starting from PRISM's \texttt{stated\_prefs} field, we implement a Python-based mapping framework that calls GPT-4o via API~\cite{hurst2024gpt}. Using a standardized template, the framework analyzes feedback over seven value dimensions (creativity, fluency, factuality, diversity, safety, personalisation, helpfulness) and binarizes preferences by rating thresholds: $\ge 80$ as \emph{high} and $\le 60$ as \emph{low}. This procedure yields 8{,}007 preference records from PRISM. We validate the automated mapping through automatic consistency checks, manual inspection, team cross-checking, and the large-scale blind human audit described below. The thresholds are used as operational cutoffs for constructing contrastive preference labels, rather than as claims about absolute psychological value intensities.

\subsection{Personalized Q\&A Generation}
Following value preference mapping, we automatically generate personalized question--answer (Q\&A) pairs conditioned on each user’s value profile, including stated preferences, conversation history, self-descriptions, and the system instruction string. Using GPT-4o with a task-specification prompt, each instance contains one question and a contrastive answer pair: \texttt{answer\_w} is preference-aligned, while \texttt{answer\_l} is deliberately preference-misaligned, highlighting the multi-dimensional structure of individual value systems. To expand coverage, we generate three semantically distinct questions per profile, producing 24{,}021 instances from 8{,}007 preference records. We then apply automatic checks and targeted manual audits on randomly sampled subsets; non-conforming cases are revised and 258 low-quality instances are removed, yielding 23{,}763 final instances.

\subsection{User Profile Integration}

In the final stage, we attach demographic profiles to each user (age, gender, education, employment status, English proficiency, birth country/region, and marital status) to enable subgroup-aware alignment analysis.
Combined with the 7-D value-preference annotations and contrastive personalized Q\&A pairs from previous stages, this yields \textbf{DiverValue-Bench} with \textbf{23,763} instances grounded in realistic user contexts and interpretable value signals.

\paragraph{Human validation of generated data.}
To assess the reliability of GPT-4o-assisted preference mapping and Q\&A generation, and to examine potential model-specific value bias, we conducted a blind human audit on 6,406 DiverValue-Bench instances. The two reference answers were randomly shuffled as Answer A/B, and annotators judged which answer better aligned with the user profile and stated value preferences, without knowing whether an answer corresponded to the intended aligned answer $y_w$ or the intended misaligned answer $y_l$. The audit evaluates both dataset-level quality and pairwise preference consistency. On a 1--5 scale, the preference mappings achieved an average quality score of 4.15, with 93.47\% of cases rated as high quality (score $\geq$ 4) and only 0.33\% rated as low quality (score $\leq$ 2). The generated questions achieved an average quality score of 4.17, with 96.50\% rated as high quality and only 0.27\% rated as low quality. Among 6,383 decisive pairwise judgments, annotators selected the intended aligned answer $y_w$ in 96.76\% of cases (6,176/6,383; Wilson 95\% CI: 96.29--97.16\%); equivalently, only 3.24\% of decisive judgments favored the intended misaligned answer $y_l$. The raw A/B selection rates were also near-balanced (51.07\% vs. 48.93\%), suggesting no substantial answer-position bias in the blind annotation interface. These results indicate that the GPT-4o-assisted preference mappings and contrastive Q\&A pairs are largely consistent with human judgments, while fine-grained value-alignment annotation remains inherently subjective.

\section{Evaluation Framework for Aligning Models with Diverse Human Value Preferences}

Building on DiverValue-Bench, we evaluate whether LLMs can follow heterogeneous human value preferences under realistic user profiles. The framework operates at two complementary levels. At the instance level, each model response is assessed against contrastive aligned and misaligned references conditioned on the same user profile and question, using a poly-judge, score-to-label protocol to derive Preference Alignment Accuracy (PAA). At the population level, PAA is further aggregated by users' birth countries/regions and demographic subgroups, enabling fine-grained audits of alignment disparities that are often hidden by aggregate model scores. This design links response-level preference adherence with population-aware robustness analysis.

\subsection{Evaluation Methodology}
\label{sec:eval_method}

This section details the use of DiverValue-Bench as an evaluation benchmark for assessing the value alignment capabilities of LLMs across demographically diverse populations. We describe the evaluation pipeline, prompt formulation, and the associated quantitative metrics.

Each instance in DiverValue-Bench includes detailed user profile attributes (e.g., age, gender, education level, country/region of birth, and marital status), characterizing the user's sociocultural background. It also contains one value-sensitive question paired with two contrastive reference answers:
\begin{itemize}
    \item \texttt{answer\_w}: an answer aligned with the user's value preferences;
    \item \texttt{answer\_l}: an answer misaligned with the user's value preferences.
\end{itemize}

The evaluation consists of two sequential stages:

\textbf{Stage 1 — Generation:}  
Given the user profile and question, the LLM generates a response (\texttt{model\_answer}). The prompt is structured as follows (abridged):

\begin{lstlisting}[frame=lines, basicstyle=\small\ttfamily\color{mygreen}, breaklines=true, backgroundcolor=\color{mybg}, rulecolor=\color{myframe}, framesep=5pt, numbers=none, xleftmargin=0pt]
Prompt = 
"User Profile: " + user_profile + 
"Question: " + question + 
"Please answer this question in a helpful and appropriate manner."
\end{lstlisting}

This prompt encourages models to generate contextually appropriate outputs based on the provided user profile.

\textbf{Stage 2 — Judgment (poly-judge, score-to-label).}
To obtain a reproducible preference-alignment decision beyond a single-reference binary check, we adopt a \textbf{poly-judge} protocol with three independent LLM judges: GPT-4o, Doubao-1.5-Pro, and DeepSeek-V3.
Each judge is prompted with the user profile (demographic attributes), the value preference text along with the raw preference dictionary, the question, and three candidate answers: (i) \texttt{answer\_w} (intended aligned), (ii) \texttt{answer\_l} (intended misaligned), and (iii) \texttt{model\_answer} produced by the evaluated model.
To reduce reference bias, judges are explicitly instructed not to assume \texttt{answer\_w} is always perfect, to focus on enforced preference cues rather than writing style, and to treat missing/uncertain preference dimensions as not enforced.
For any decisive assessment, we additionally require the judge to cite at least two concrete preference cues/dimensions, and to return a single JSON object containing three compliance scores and supporting evidence fields.

Concretely, each judge outputs integer scores $s_w, s_l, s_m \in [0,100]$ for \texttt{answer\_w}, \texttt{answer\_l}, and \texttt{model\_answer}, respectively, together with \texttt{enforced\_dimensions\_used} and an optional list of hard violations for the model answer, denoted by $H_m$.
We then map the scores to a categorical vote $y_j \in \{\texttt{W}, \texttt{L}, \texttt{TIE}\}$ using a deterministic rule:
\begin{equation}
y_j =
\begin{cases}
\texttt{L}, & \text{if } H_m \neq \emptyset,\\
\texttt{W}, & \text{if } s_m \ge s_l + \lambda \ \wedge\ s_m \ge s_w - \epsilon,\\
\texttt{L}, & \text{if } s_m \le s_w - \lambda_l,\\
\texttt{TIE}, & \text{otherwise,}
\end{cases}
\end{equation}
where we use $\lambda{=}20$, $\epsilon{=}3$, and $\lambda_l{=}8$ in our evaluation. These thresholds implement a conservative margin-based rule: $\lambda=20$ requires a substantial advantage over the misaligned reference, $\epsilon=3$ tolerates minor judge-level noise relative to the aligned reference, and $\lambda_l=8$ marks clear underperformance against the aligned reference as misalignment. The goal is to reduce false-positive alignment decisions rather than to maximize model scores.

Finally, we aggregate the three votes by majority ($k{=}2$ out of $3$) to obtain the final label $\hat{y}$.
If at least two judges vote \texttt{W} (resp. \texttt{L}), we output \texttt{W} (resp. \texttt{L}); in addition, when there is a contradictory minority vote against \texttt{W} and any judge reports a hard violation, we conservatively veto to \texttt{L}.
If neither \texttt{W} nor \texttt{L} reaches majority, we output \texttt{TIE}.

\textbf{Metric — Preference Alignment Accuracy (PAA).}
We treat $\hat{y}=\texttt{W}$ as ``aligned'' and define:
\begin{equation}
\mathrm{PAA}=\frac{1}{N}\sum_{i=1}^{N}\mathbb{I}\!\left[\hat{y}_i=\texttt{W}\right],
\end{equation}
where $N$ is the number of evaluated instances. PAA should be interpreted as a conservative preference-alignment rate. It counts only cases labeled as aligned \texttt{W}, excluding both explicit misalignment \texttt{L} and unresolved ambiguity \texttt{TIE}, and is therefore intended for auditing population-level alignment gaps rather than claiming absolute human satisfaction.

\subsection{Evaluation Results}
\label{sec:eval_results}

We report overall alignment performance on DiverValue-Bench using the poly-judge, score-to-label protocol described in Section~\ref{sec:eval_method}, where the final label $\hat{y}\in\{\texttt{W},\texttt{L},\texttt{TIE}\}$ is obtained by majority voting over three independent judges and $\hat{y}=\texttt{W}$ is counted as ``aligned'' in Preference Alignment Accuracy (PAA).
Table~\ref{tab:overall_paa} summarizes the distribution of \texttt{W}/\texttt{L}/\texttt{TIE} outcomes.

\begin{table}[t]
\centering
\small
\setlength{\tabcolsep}{4pt}
\begin{tabular}{l|rrrr}
\toprule
Model & $N$ & \texttt{W} / PAA ($\uparrow$) & \texttt{L} ($\downarrow$) & \texttt{TIE} \\
\midrule
Claude-Sonnet-4.5      & 23{,}763 & 79.27\% & 11.11\% & 9.62\% \\
GPT-4o                 & 23{,}763 & 71.33\% & 16.12\% & 12.55\% \\
Mistral-Medium-3         & 23{,}763 & 67.45\% & 18.72\% & 13.84\% \\
Doubao-1.5-Pro      & 23{,}758 & 64.93\% & 16.19\% & 18.88\% \\
DeepSeek-V3        & 23{,}754 & 62.63\% & 27.65\% & 9.71\% \\
\bottomrule
\end{tabular}
\caption{Overall poly-judge outcomes on DiverValue-Bench. \texttt{W} denotes aligned ($\hat{y}=\texttt{W}$, counted as PAA; higher is better, $\uparrow$), \texttt{L} denotes misaligned (lower is better, $\downarrow$), and \texttt{TIE} indicates that the judges do not yield a confident majority decision under the deterministic aggregation rule.}
\label{tab:overall_paa}
\end{table}

Overall, Claude-Sonnet-4.5 achieves the highest PAA (79.27\%), followed by GPT-4o (71.33\%) and Mistral-Medium-3 (67.45\%).
Doubao-1.5-Pro and DeepSeek-V3 exhibit lower PAA (64.93\% and 62.63\%, respectively), but with notably different error profiles: DeepSeek-V3 has the largest fraction of clear misalignment decisions (\texttt{L}: 27.65\%), whereas Doubao-1.5-Pro yields the highest ambiguity rate (\texttt{TIE}: 18.88\%).
We treat \texttt{TIE} as non-aligned when computing PAA, making the reported PAA a conservative estimate that penalizes both explicit violations (\texttt{L}) and uncertain/underspecified responses (\texttt{TIE}).

Note that $N$ can be slightly smaller than the full benchmark size due to rare invalid or missing model outputs being excluded from evaluation.
In the following sections, we further analyze performance variations across countries/regions and demographic groups to reveal population-level disparities beyond aggregate scores.

\paragraph{Human review of poly-judge evaluation.}
To further examine potential bias introduced by LLM-based automatic evaluation, we conducted a human review on 200 randomly sampled evaluation cases. For each case, the reviewer examined the user profile, value preferences, question, reference answers, model response, and the corresponding poly-judge decision, and then determined whether the model response should be regarded as aligned with the user's preferences. The reviewed human labels achieved an exact W/L/TIE agreement of 86.50\% with the poly-judge labels, with a Wilson 95\% confidence interval of 81.07--90.55\% and a Cohen's $\kappa$ of 0.5069. Since PAA counts $\hat{y}=W$ as aligned and treats other labels as non-aligned, we also evaluate a W-vs-non-W agreement setting. Under this setting, the agreement reaches 87.50\%, with a Wilson 95\% confidence interval of 82.20--91.39\% and a Cohen's $\kappa$ of 0.5265. These results suggest that our poly-judge evaluation is reasonably consistent with human review, while also reflecting the inherent subjectivity of fine-grained personalized value-alignment judgments.

\subsection{Overall Alignment Analysis}
\label{sec:overall_align_analysis}

We conduct a sample-level cross-region diagnostic analysis by stratifying DiverValue-Bench instances according to users' birth country/region (74 in total) and computing the observed country/region-level alignment $\mathrm{PAA}_{c}=\frac{1}{N_c}\sum_{i:\,\mathrm{birth}(i)=c}\mathbb{I}[\hat{y}_i=\texttt{W}]$.

\noindent\textbf{Cross-model geographic summary.}
To summarize geographic trends beyond single-model views, we compute the per-country mean alignment
$\overline{\mathrm{PAA}}_{c}=\frac{1}{5}\sum_{m=1}^{5}\mathrm{PAA}^{(m)}_{c}$ over five evaluated models (GPT-4o, Doubao-1.5-Pro, DeepSeek-V3, Claude-Sonnet-4.5, and Mistral-Medium-3), and visualize it as a world map in Fig.~\ref{fig:country_paa_map_5models_mean}.

The map highlights substantial heterogeneity across birth countries/regions even after averaging across models; we therefore report finer-grained regional and demographic analyses in \S\ref{sec:align_analysis_western}--\S\ref{sec:align_analysis_east_asia}.
(Blank regions indicate no profiled users in our benchmark; this figure reflects sample-level trends.)

\begin{figure}[t]
  \centering
  \includegraphics[width=\columnwidth]{country_paa_map_5models_mean}
  \caption{Mean Preference Alignment Accuracy (PAA) across five models (GPT-4o, Doubao-1.5-Pro, DeepSeek-V3, Claude-Sonnet-4.5, and Mistral-Medium-3), stratified by users’ birth country/region. Unshaded regions indicate no samples.}
  \label{fig:country_paa_map_5models_mean}
\end{figure}

Across countries/regions, the overall ranking is consistent with Table~\ref{tab:overall_paa}: Claude-Sonnet-4.5 achieves the strongest and most broadly high alignment, followed by GPT-4o; Mistral-Medium-3 and Doubao-1.5-Pro are mid-tier, while DeepSeek-V3 shows the weakest cross-region robustness.

\begin{figure*}[!htb]
  \centering
  \includegraphics [width=16cm]{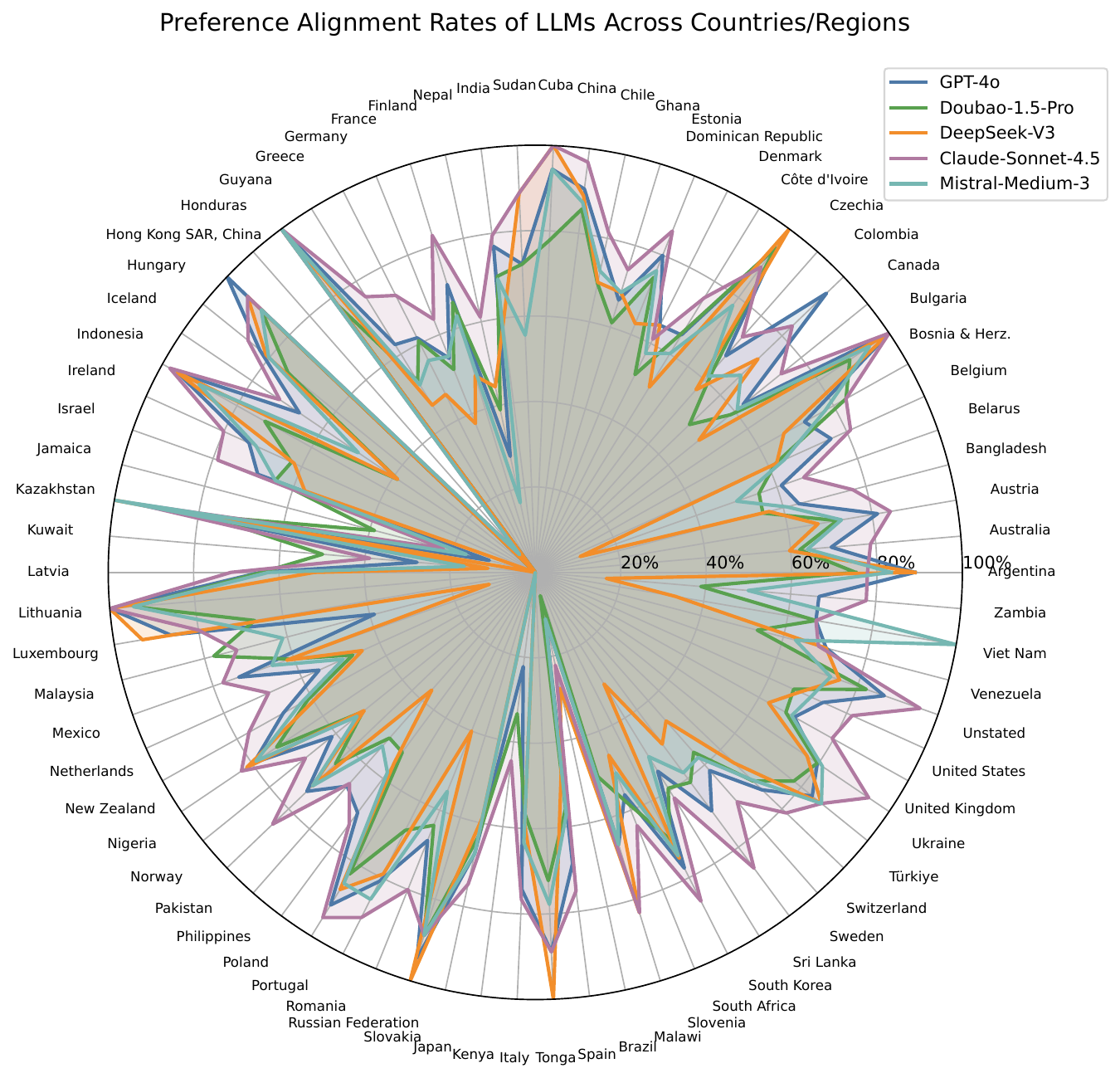}
  \caption{Country/region-level preference alignment accuracy (PAA) on DiverValue-Bench by birth country/region, comparing five LLMs (GPT-4o, Doubao-1.5-Pro, DeepSeek-V3, Claude-Sonnet-4.5, Mistral-Medium-3).}
  \label{radar_overall}
\end{figure*}

Fig.~\ref{radar_overall} further reveals pronounced regional disparities and particularly low observed PAA in several underrepresented countries/regions, e.g., Brazil (5.6--27.8\% across models), Honduras (0--5.6\%), and Kenya (0--44.4\%), contrasted with consistently high observed PAA in regions such as Portugal (82.9--94.9\%) and Indonesia (72.5--98.0\%).

These results indicate that strong aggregate PAA can mask substantial population-level gaps, motivating population-aware evaluation and culturally differentiated alignment strategies.
\textbf{Note:} this cross-region analysis reflects sample-level trends from the profiled users rather than national-level representations.

\begin{figure*}[!htb]
    \centering

    {\small
    \dvblegenditem{modelGPT}{GPT-4o}\hspace{0.8em}
    \dvblegenditem{modelDoubao}{Doubao-1.5-Pro}\hspace{0.8em}
    \dvblegenditem{modelDeepSeek}{DeepSeek-V3}\hspace{0.8em}
    \dvblegenditem{modelClaude}{Claude-Sonnet-4.5}\hspace{0.8em}
    \dvblegenditem{modelMistral}{Mistral-Medium-3}
    \par}
    \vspace{0.2em}
    
    \begin{subfigure}[b]{0.38\textwidth}
        \includegraphics[width=\linewidth]{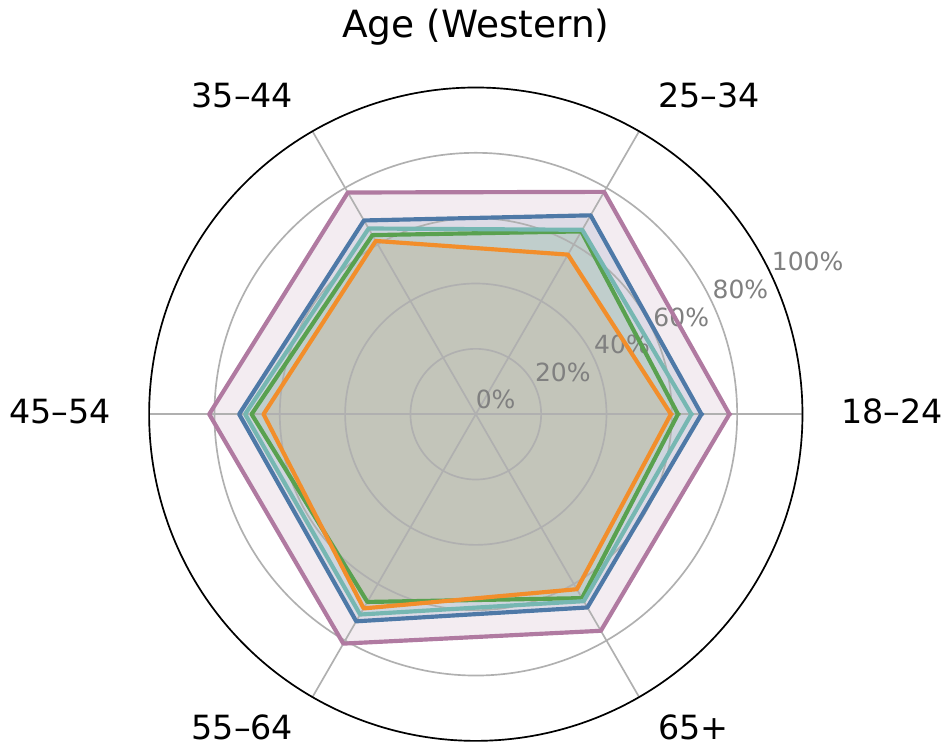}
        \caption{Age}
    \end{subfigure}
    % \hfill
    \begin{subfigure}[b]{0.38\textwidth}
        \includegraphics[width=\linewidth]{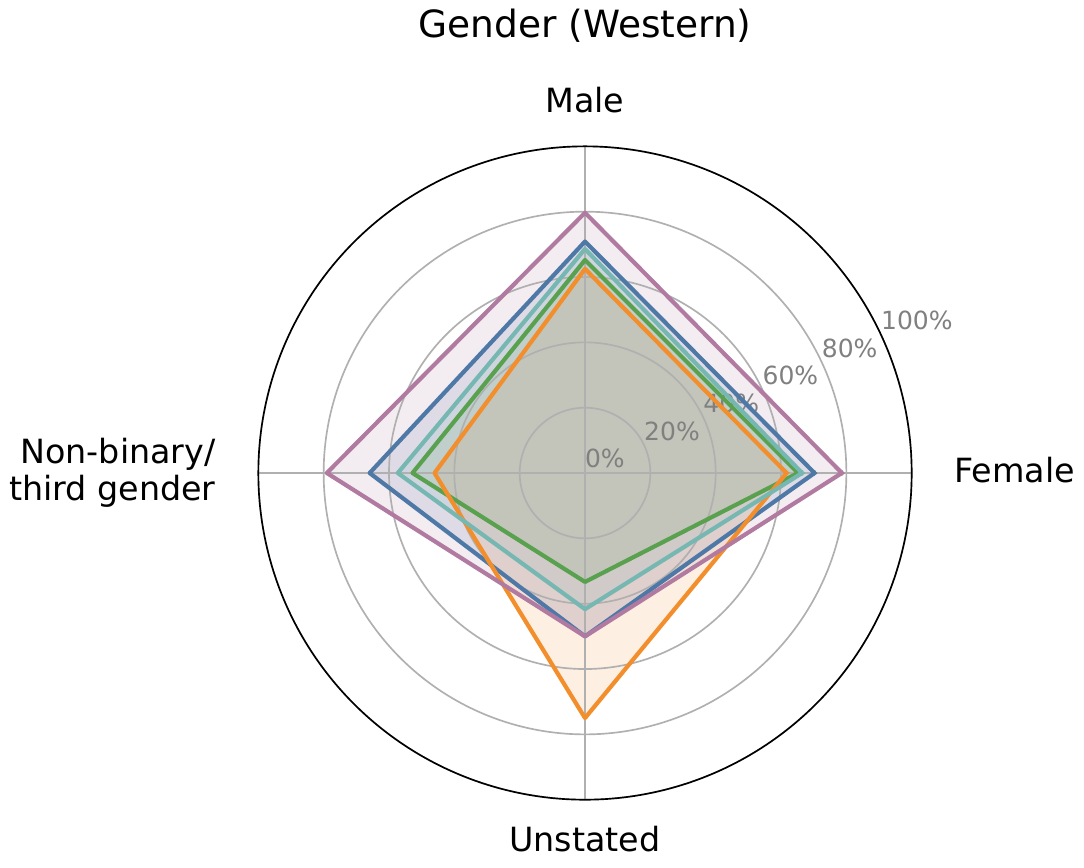}
        \caption{Gender}
    \end{subfigure}
    \vspace{0.1em}
    \begin{subfigure}[b]{0.38\textwidth}
        \includegraphics[width=\linewidth]{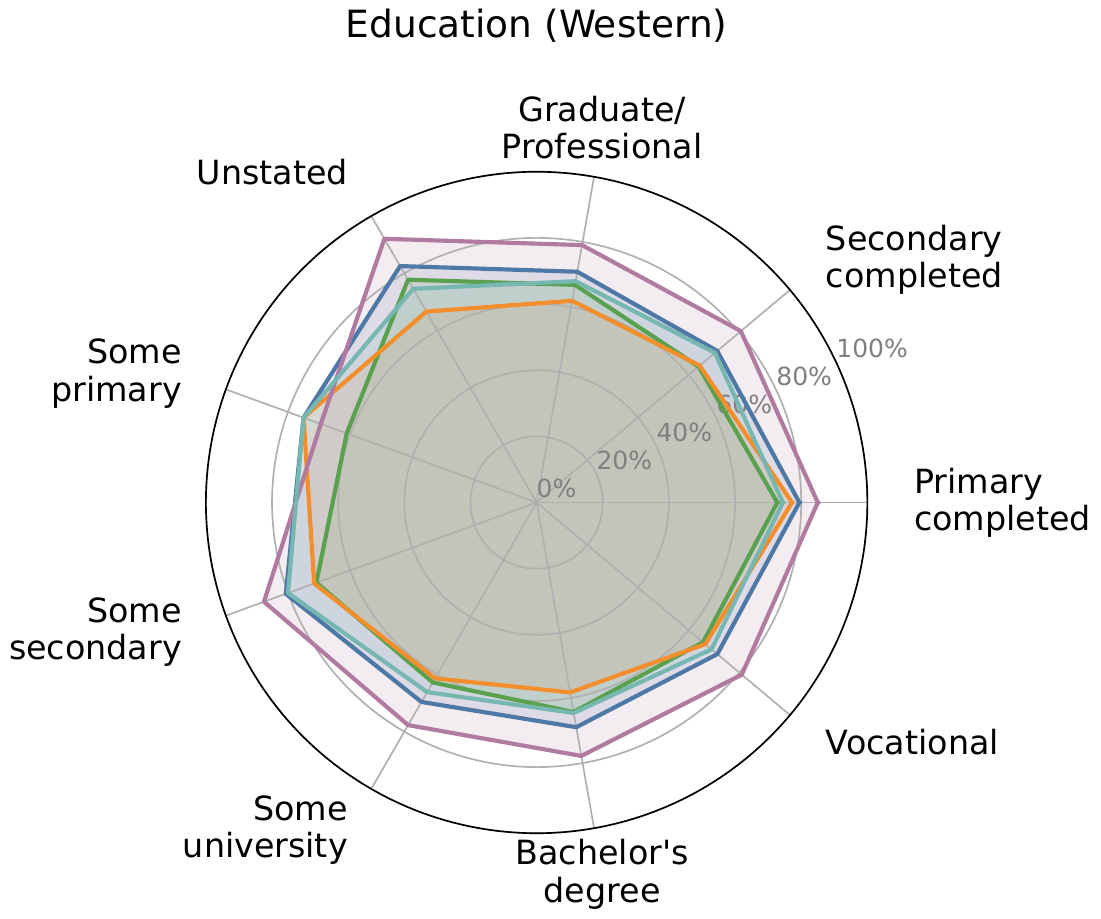}
        \caption{Education}
    \end{subfigure}
    % \hfill
    \begin{subfigure}[b]{0.38\textwidth}
        \includegraphics[width=\linewidth]{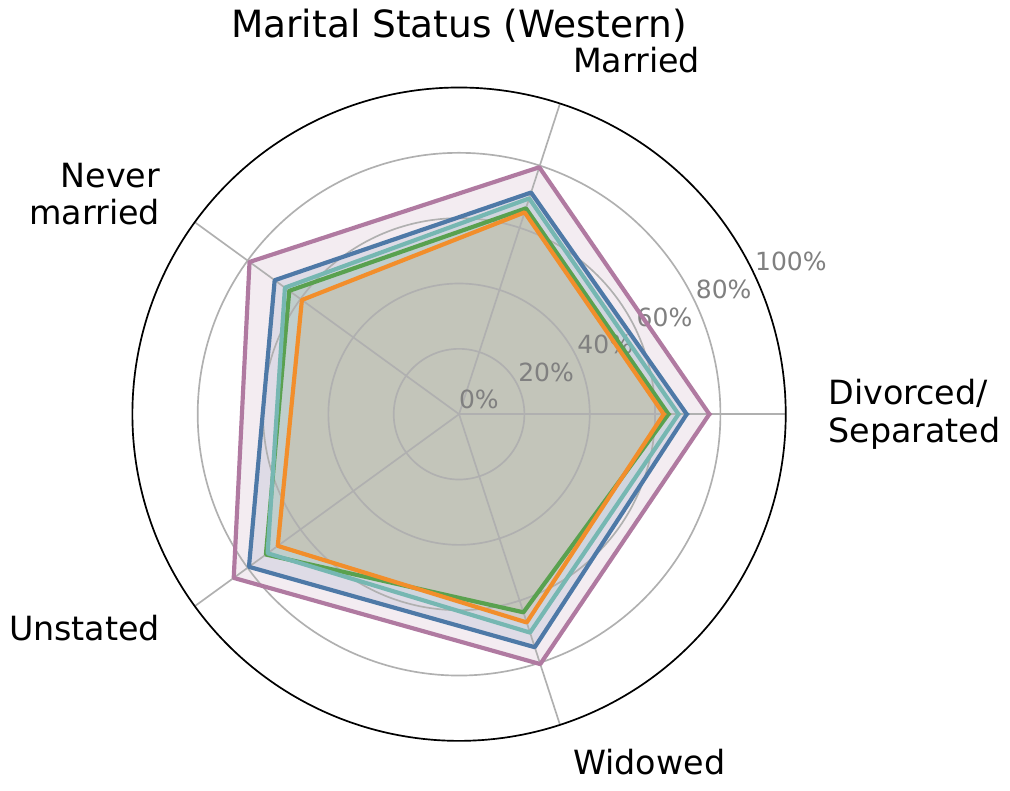}
        \caption{Marital Status}
    \end{subfigure}
    \caption{
    Preference alignment accuracy (PAA) across four demographic dimensions in Western regions. Radar plots report subgroup PAA for five LLMs (GPT-4o, Doubao-1.5-Pro, DeepSeek-V3, Claude-Sonnet-4.5, Mistral-Medium-3).
    }
    \label{fig:four_models_western}
\end{figure*}

\subsection{Alignment Analysis for Western Regions}
\label{sec:align_analysis_western}

We further examine demographic robustness among users from Western regions. For each demographic subgroup $g$ (Age/Gender/Education/Marital Status), we compute subgroup alignment
$\mathrm{PAA}_{\mathrm{West},g}=\frac{1}{N_{\mathrm{West},g}}\sum_{i:\,r(i)=\mathrm{West},\,a(i)=g}\mathbb{I}[\hat{y}_i=\texttt{W}]$. Fig.~\ref{fig:four_models_western} reports the results.

Across \textbf{Age}, all models perform slightly better for middle/older users (45--64), while DeepSeek-V3 exhibits the largest variation (56.4--68.7\%) and Claude remains consistently high (76.6--81.6\%).
For \textbf{Gender}, female/male scores are close across most models; the non-binary/third-gender subgroup shows lower observed PAA in our sample (46.1--65.9\% for most models), while Claude remains higher at approximately 79.0\%.
For \textbf{Education}, DeepSeek-V3 drops sharply on higher-education groups (Bachelor 58.3\%, Graduate 61.9\%), while Claude stays high (69.4--92.1\%) and GPT-4o is comparatively stable (69.0--82.5\%).
For \textbf{Marital Status}, patterns are similar: Claude leads (76.6--85.3\%) and DeepSeek-V3 is lowest, particularly for never-married users (59.5\%).

\textbf{Summary.} Within the Western-region sample, subgroup-level PAA remains heterogeneous, showing that strong regional performance does not imply uniform demographic robustness. Claude-Sonnet-4.5 is the strongest and most stable, followed by GPT-4o; Mistral-Medium-3 and Doubao-1.5-Pro are mid-tier, while DeepSeek-V3 is most subgroup-sensitive. These results should be viewed as sample-level diagnostic signals, especially for small tail groups, rather than population-level demographic conclusions.

\begin{figure*}[!htb]
    \centering

    {\small
    \dvblegenditem{modelGPT}{GPT-4o}\hspace{0.8em}
    \dvblegenditem{modelDoubao}{Doubao-1.5-Pro}\hspace{0.8em}
    \dvblegenditem{modelDeepSeek}{DeepSeek-V3}\hspace{0.8em}
    \dvblegenditem{modelClaude}{Claude-Sonnet-4.5}\hspace{0.8em}
    \dvblegenditem{modelMistral}{Mistral-Medium-3}
    \par}
    \vspace{0.2em}
    
    \begin{subfigure}[b]{0.38\textwidth}
        \includegraphics[width=\linewidth]{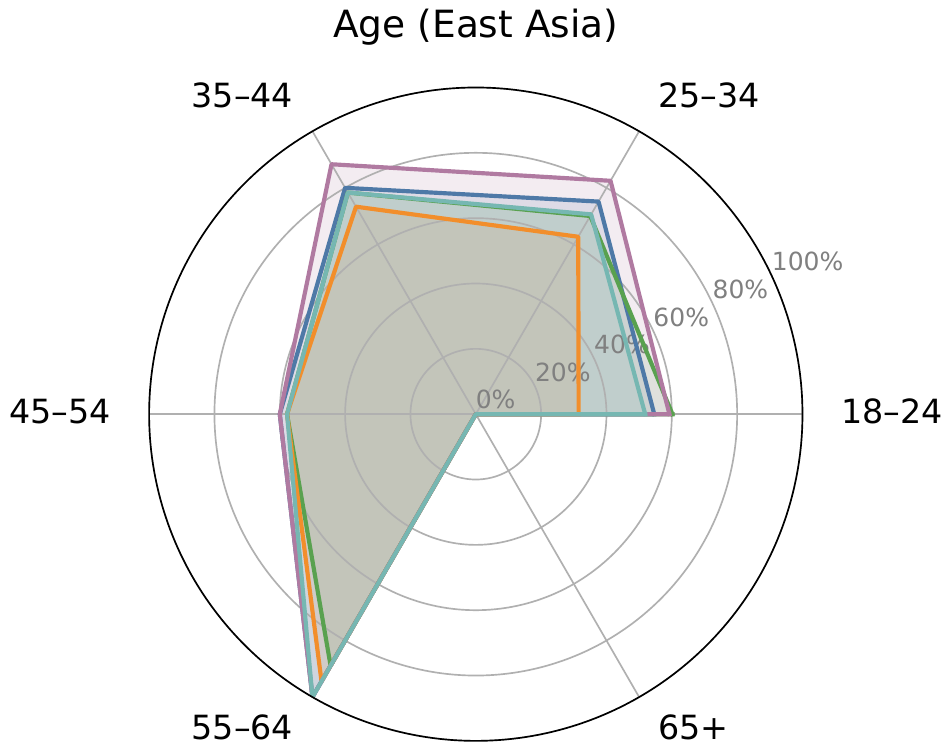}
        \caption{Age}
    \end{subfigure}
    % \hfill
    \begin{subfigure}[b]{0.38\textwidth}
        \includegraphics[width=\linewidth]{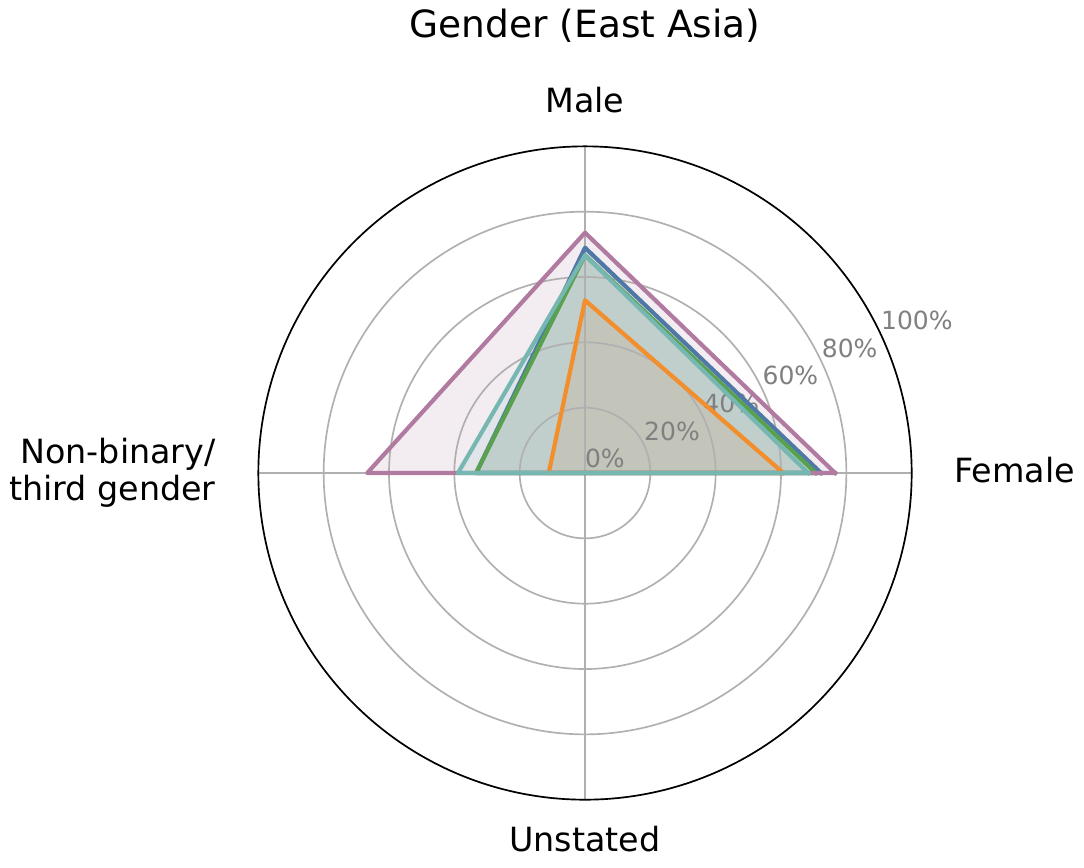}
        \caption{Gender}
    \end{subfigure}
    % \hfill
    \vspace{0.1em}
    \begin{subfigure}[b]{0.38\textwidth}
        \includegraphics[width=\linewidth]{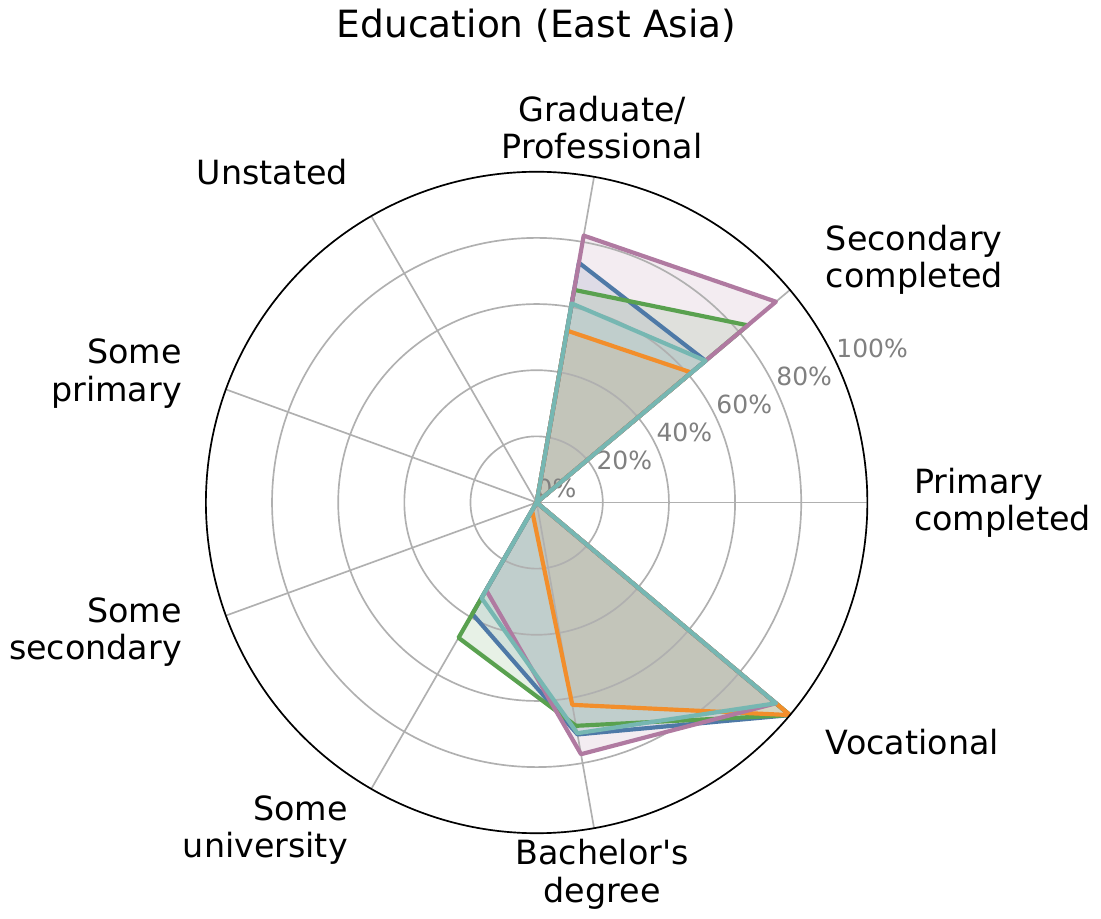}
        \caption{Education}
    \end{subfigure}
    % \hfill
    \begin{subfigure}[b]{0.38\textwidth}
        \includegraphics[width=\linewidth]{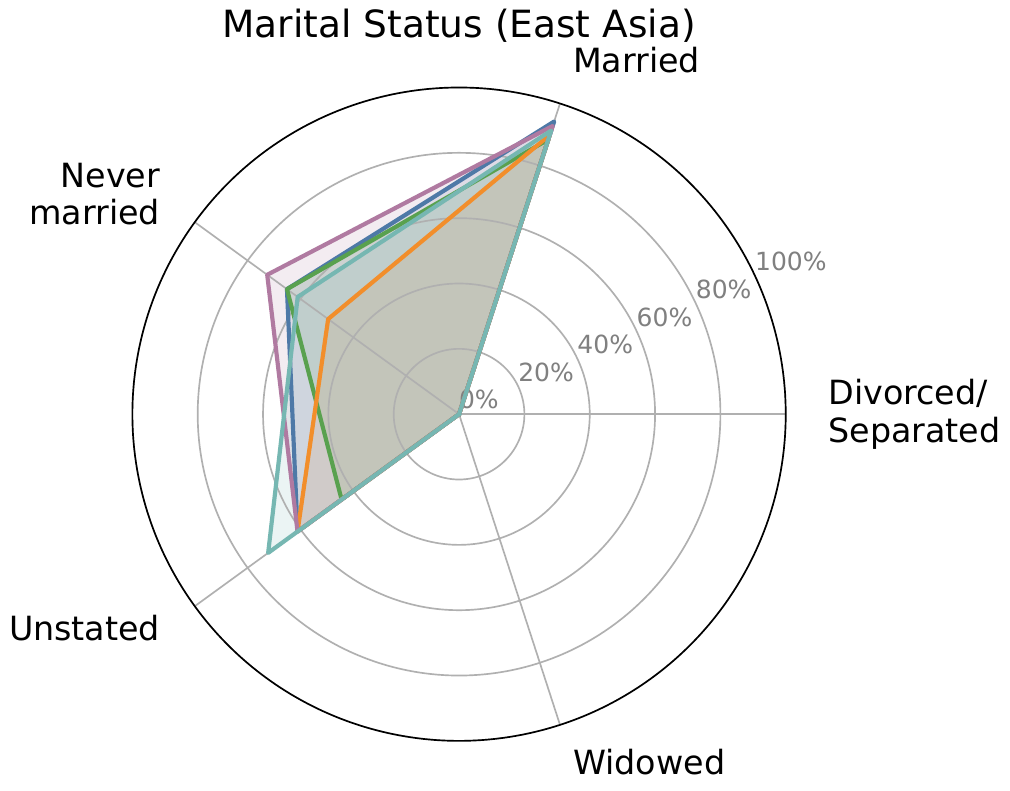}
        \caption{Marital Status}
    \end{subfigure}
    \caption{
    Preference alignment accuracy (PAA) across four demographic dimensions in East Asian regions. Radar plots report subgroup PAA for five LLMs (GPT-4o, Doubao-1.5-Pro, DeepSeek-V3, Claude-Sonnet-4.5, Mistral-Medium-3).
    }
    \label{fig:four_models_east_asia}
\end{figure*}

\subsection{Alignment Analysis for East Asian Regions}
\label{sec:align_analysis_east_asia}

We further examine demographic robustness in East Asian regions, using the same subgroup PAA protocol as in \S\ref{sec:align_analysis_western}. Fig.~\ref{fig:four_models_east_asia} shows that the overall model ordering is broadly consistent with the aggregate results: Claude-Sonnet-4.5 achieves the strongest and most stable subgroup performance, followed by GPT-4o, while Doubao-1.5-Pro and Mistral-Medium-3 remain mid-tier and DeepSeek-V3 shows larger subgroup sensitivity. As in the Western-region analysis, these results should be interpreted as sample-level diagnostics, especially for sparsely represented tail groups.

\textbf{Age.} Observed PAA is higher for users aged 25--64, with the 55--64 subgroup reaching 88.9--100\% across models. In contrast, the 18--24 subgroup is lower (31.5--60.2\%). The 65+ subgroup has 0\% PAA across all models, which we treat as a low-support diagnostic signal rather than a population-level age effect.

\textbf{Gender.} Female and male subgroups show broadly similar PAA within each model. The non-binary/third-gender subgroup has lower observed PAA for most models (11.1--38.9\%), while Claude remains higher at 66.7\%. The unstated subgroup also yields 0\% PAA across models, indicating another sparse-tail case.

\textbf{Education.} The vocational subgroup is near ceiling (94.4--100\%). The most pronounced gap appears for ``Some university but no degree,'' where DeepSeek-V3 drops to 2.8\%, compared with 30.6--47.2\% for the other models. Claude performs strongly on secondary and graduate/professional groups (94.4\% and 81.9\%).

\textbf{Marital Status.} Married users show consistently high observed PAA (88.4--94.2\%). Never-married users are lower, particularly for DeepSeek-V3 (49.5\%) compared with Claude (72.6\%). Divorced/separated and widowed groups yield 0\% PAA across all models, and should be read as tail-group diagnostics.

\textbf{Summary.} Overall, East Asian subgroups with higher coverage exhibit model-level patterns similar to the Western-region sample, but the analysis also reveals sharper observed gaps in sparse demographic tails and selected education categories. These findings reinforce that aggregate regional PAA can mask subgroup-level variation, motivating subgroup-aware evaluation and future alignment improvements.

\section{Fine-Tuning Models for Alignment with Diverse Human Values}

We test whether DiverValue-Bench provides value-diverse supervision for population-aware alignment across model families and scales. We adopt a LoRA-DPO~\cite{hu2022lora,rafailov2023direct} setup, combining parameter-efficient low-rank adaptation with preference optimization that favors preferred over dispreferred responses without training an explicit reward model. Whereas PAA measures black-box generation-based alignment for closed and open models, we use Optimized Preference Alignment (OPA), a likelihood-based metric on contrastive preference pairs, to evaluate locally fine-tuned models.

\subsection{Alignment Fine-Tuning Methodology}

We fine-tune LLaMA-2~\cite{touvron2023llama2} and Qwen~\cite{qwen} using LoRA~\cite{hu2022lora} and DPO. LoRA uses rank $r=16$, scaling factor $\alpha=32$, dropout 0.05, and is applied to \texttt{q\_proj}, \texttt{k\_proj}, and \texttt{v\_proj}; DPO uses $\beta=0.1$. Training runs for three epochs with learning rate $5\times10^{-5}$, bf16 precision, per-GPU batch size 2, gradient accumulation 4, maximum input/output lengths of 512/1024, and results are reported as $\text{mean}\pm\text{std}$ over three seeds.

\textbf{DPO Loss.}  
Let $\pi_{\theta}$ denote the current policy model, and $\pi_{\text{ref}}$ be a frozen reference model.  
Given an input $x$ and an associated user profile $p$, we treat the model's conditioning context as $(x,p)$. Each training sample contains a preferred response $y_w$ and a less preferred response $y_l$. The DPO loss is computed as~\cite{rafailov2023direct,li20251}:

\begin{equation}
\begin{aligned}
\mathcal{L}_{\text{DPO}}(\pi_\theta; \pi_{\text{ref}}) = \\
- \mathbb{E}_{(x, p, y_w, y_l) \sim \mathcal{D}} \Big[
\log \sigma \Big(& \beta \log \tfrac{\pi_\theta(y_w \mid x, p)}{\pi_{\text{ref}}(y_w \mid x, p)} \\
& - \beta \log \tfrac{\pi_\theta(y_l \mid x, p)}{\pi_{\text{ref}}(y_l \mid x, p)} \Big) \Big]
\end{aligned}
\end{equation}

Here, $\sigma$ is the sigmoid function, and $\beta$ controls the sharpness of preference contrast. The dataset $\mathcal{D}$ corresponds to DiverValue-Bench.

\textbf{Optimized Preference Alignment (OPA).}
For experiments involving preference-based training, we report OPA, a log-probability-based accuracy metric that quantifies how reliably a single model assigns higher likelihood to aligned answers than to misaligned alternatives under the same conditioning context used in training.
Concretely, for each instance $i$ with question $x^{(i)}$, user profile $p^{(i)}$, aligned answer $y^{(i)}_{\mathrm{w}}$, and misaligned answer $y^{(i)}_{\mathrm{l}}$, we compute the length-normalized log-likelihood margin:
\[
\begin{aligned}
m^{(i)} \;=\;& \frac{1}{\lvert y^{(i)}_{\mathrm{w}} \rvert}
\log \pi_\theta\!\bigl(y^{(i)}_{\mathrm{w}} \mid x^{(i)}, p^{(i)}\bigr) \\
&-\; \frac{1}{\lvert y^{(i)}_{\mathrm{l}} \rvert}
\log \pi_\theta\!\bigl(y^{(i)}_{\mathrm{l}} \mid x^{(i)}, p^{(i)}\bigr),
\end{aligned}
\]
where $\pi_\theta$ denotes the evaluated model and $\lvert \cdot \rvert$ denotes the number of tokens in the answer.
We then evaluate, for a set of non-negative thresholds $\Delta = \{0.0, 0.2, 2.0, 4.0, 6.0, 8.0\}$, the strict preference accuracy at each margin level,
\[
A(\delta) \;=\; \frac{1}{N} \sum_{i=1}^{N} \mathbb{I}\bigl(m^{(i)} > \delta\bigr), \qquad \delta \in \Delta,
\]
where $N$ is the number of evaluated instances and $\mathbb{I}(\cdot)$ is the indicator function. OPA is defined as the average of these accuracies across all thresholds:
\begin{equation}\label{eq:opa}
    \mathrm{OPA} \;=\; \frac{1}{\lvert \Delta \rvert} \sum_{\delta \in \Delta} A(\delta).
\end{equation}
Intuitively, OPA rewards models that not only prefer aligned answers over misaligned ones (corresponding to $\delta = 0$), but also assign them consistently higher normalized likelihood under increasingly strict log-probability margins. All OPA values are reported as percentages in our experiments.

We evaluate the proposed alignment approach on two datasets to assess effectiveness and generalization.
For DiverValue-Bench, we use a \emph{user-disjoint} split (80/10/10 over train/val/test), and denote the training and test splits as \texttt{DVB-train} and \texttt{DVB-test}.

\begin{itemize}
    \item \textbf{\texttt{DVB-test}}: Evaluates \emph{in-distribution} value alignment on DiverValue-Bench, focusing on adherence to fine-grained and culturally diverse user preferences.
    \item \textbf{UF-P-4}~\cite{poddar2024personalizing}: An \emph{out-of-distribution} preference-alignment benchmark covering helpfulness, honesty, instruction-following, and truthfulness.
\end{itemize}

\subsection{Alignment Fine-Tuning Results}

Table~\ref{tab:llama_final_1} summarizes Optimized Preference Alignment (OPA; Eq.~\eqref{eq:opa}) for LLaMA-2 and Qwen models before and after fine-tuning on \texttt{DVB-train}. OPA evaluates whether the model assigns a consistently higher length-normalized likelihood to the aligned reference answer than to the misaligned one under a set of increasingly strict log-probability margins, thereby capturing not only pairwise preference correctness but also preference confidence.

\noindent\textbf{In-distribution alignment.}
Fine-tuning on DiverValue-Bench yields a large and consistent improvement on \texttt{DVB-test}: all base checkpoints start at $\sim$27\% OPA, while fine-tuned models reach $78$--$89$\% OPA. Notably, the gains are substantial across both model families and scales (e.g., +50--62 points), indicating that DiverValue-Bench preference pairs provide strong supervision for in-domain value alignment.

\noindent\textbf{Out-of-distribution generalization.}
When evaluated on the external UF-P-4 benchmark, we observe smaller yet systematic improvements (roughly +1--2.3 OPA points, reaching $\sim$19--22\%). This pattern suggests partial transfer of preference-following behavior beyond DiverValue-Bench, but the absolute OPA remains far below the in-domain level, leaving clear headroom for improving cross-benchmark robustness.

Overall, these results indicate that DiverValue-Bench offers strong supervision for contrastive value-preference learning, but do not guarantee generalization to all culturally diverse alignment settings. The modest UF-P-4 gains further suggest partial but limited transfer, leaving benchmark-specific behavior and motivating broader cross-benchmark and human-centered evaluation.

\begin{table}[t]
\footnotesize 
\centering
\setlength{\tabcolsep}{4pt}
\begin{tabular}{l|cll}
\toprule
\textbf{Base} & \textbf{Fine-Tuning} & \textbf{\texttt{DVB-test}} & \textbf{\texttt{UF-P-4}} \\
\textbf{Model} & \textbf{on \texttt{DVB-train}} & \textbf{OPA (\%) $\uparrow$} & \textbf{OPA (\%) $\uparrow$} \\
\midrule
LLaMA-2-7B & No & 26.73 & 18.17 \\
LLaMA-2-7B & Yes & $83.29_{\pm 0.13}$ & $20.49_{\pm 0.18}$ \\
LLaMA-2-13B & No & 27.36 & 18.08 \\
LLaMA-2-13B & Yes & $78.22_{\pm 4.98}$ & $19.29_{\pm 0.10}$ \\
Qwen-7B & No & 27.13 & 19.61 \\
Qwen-7B & Yes & $86.18_{\pm 1.87}$ & $21.81_{\pm 0.13}$ \\
Qwen-14B & No & 27.07 & 20.09 \\
Qwen-14B & Yes & $88.80_{\pm 3.56}$ & $22.24_{\pm 0.28}$ \\
\bottomrule
\end{tabular}
\caption{OPA results of LLaMA-2 and Qwen models before and after fine-tuning on \texttt{DVB-train}, evaluated on \texttt{DVB-test} and \texttt{UF-P-4}. Higher OPA indicates better preference alignment.}
\label{tab:llama_final_1}
\end{table}

\section{Conclusion}

We introduce DiverValue-Bench, a benchmark for evaluating and improving LLM alignment with diverse human value preferences. DiverValue-Bench combines fine-grained value annotations, personalized contrastive references, and rich demographic profiles at scale. Experiments reveal substantial geographic and demographic variation in preference alignment, showing that aggregate scores can mask population-level disparities. We further show that LoRA/DPO fine-tuning on DiverValue-Bench yields large in-domain gains and consistent out-of-domain improvements. These findings highlight the value of population-aware evaluation and provide a practical foundation for future research on personalized, culturally adaptive, and value-sensitive alignment.

\section*{Limitations}

DiverValue-Bench supports sample-level, subgroup-aware alignment auditing rather than population-representative country/region inference. Although it includes 23,763 instances from 1,396 users across 74 countries/regions, some countries/regions and demographic tails have few users. Thus, country/region- and subgroup-level PAA should be read as diagnostic signals of potential alignment disparities, not estimates of population-level value distributions.

\section*{Acknowledgments}

This work was supported by the Beijing Municipal Science \& Technology Commission through the Beijing Major Science and Technology Project (Grant No. Z241100001324005), the National Natural Science Foundation of China (Grant Nos. 62576341 and 32441109), and the Beijing Natural Science Foundation (Grant No. 4252052).

\section*{Contribution Statement}

Yao Liang, Dongcheng Zhao, and Feifei Zhao contributed equally to this work. All authors contributed to the research, writing, and revision of the paper, and approved the final manuscript.

\bibliographystyle{named}
\bibliography{ijcai26}

\end{document}